\documentclass[10pt,twocolumn,letterpaper]{article}

\usepackage[algorithms]{wacv}      % To produce the REVIEW version for the algorithms track
\usepackage{graphicx}
\usepackage{amsmath}
\usepackage{amssymb}
\usepackage{booktabs}
\usepackage[linesnumbered,ruled,vlined]{algorithm2e}
\usepackage{times}
\usepackage{comment}
\usepackage{epsfig}
\usepackage{graphicx}
\usepackage{amsmath}
\usepackage{amssymb}
\usepackage{url}
\usepackage{mathrsfs}
\usepackage{multirow}
\usepackage{listings}
\usepackage{makecell}
\usepackage{colortbl}
\usepackage{adjustbox}
\usepackage{booktabs}
\usepackage{array}
\usepackage{newfloat}
\usepackage[dvipsnames]{xcolor}
\definecolor{lightgray}{rgb}{0.9, 0.9, 0.9}
\definecolor{ballblue}{rgb}{0.13, 0.67, 0.8}
\definecolor{lightgreen}{rgb}{0.95, 0.95, 0.95}
\definecolor{gg}{HTML}{e2f0cb}

\definecolor{codegreen}{rgb}{0,0.6,0}
\definecolor{codegray}{rgb}{0.5,0.5,0.5}
\definecolor{codepurple}{rgb}{0.58,0,0.82}
\definecolor{backcolour}{rgb}{0.95,0.95,0.92}

\lstdefinestyle{mystyle}{
    backgroundcolor=\color{backcolour},   
    commentstyle=\color{codegreen},
    keywordstyle=\color{magenta},
    numberstyle=\tiny\color{codegray},
    stringstyle=\color{codepurple},
    basicstyle=\ttfamily\footnotesize,
    breakatwhitespace=false,         
    breaklines=true,                 
    captionpos=b,                    
    keepspaces=true,                 
    numbers=left,                    
    numbersep=5pt,                  
    showspaces=false,                
    showstringspaces=false,
    showtabs=false,                  
    tabsize=2
}

\usepackage[pagebackref,breaklinks=true,colorlinks,bookmarks=false,citecolor=ballblue]{hyperref}

\newcommand{\STAB}[1]{\begin{tabular}{@{}c@{}}#1\end{tabular}}

\usepackage[capitalize]{cleveref}
\crefname{section}{Sec.}{Secs.}
\Crefname{section}{Section}{Sections}
\Crefname{table}{Table}{Tables}
\crefname{table}{Tab.}{Tabs.}

\begin{document}

\title{Rethinking Low-Rank Adaptation in Vision: Exploring Head-Level Responsiveness across Diverse Tasks}

\author{Yibo Zhong\\
Sichuan University\\
{\tt\small zhongyibo@stu.scu.edu.cn}
\and
Yao Zhou\\
Sichuan University\\
{\tt\small yaozhou@scu.edu.cn}
}
\maketitle

%%%%%%%%% ABSTRACT
\begin{abstract}
Low-rank adaptation (LoRA) has shifted the paradigm of adapting pre-trained Vision Transformers (ViT), achieving great efficiency by updating only a subset of tailored parameters to approximate weight updates. However, the multi-head design of the self-attention mechanism, with the heads working in parallel in the computation flow, exhibiting similar visual patterns and requiring update over all of them, incurs unnecessary storage and computational overhead. In this paper, we propose \textbf{Hea}d-level \textbf{r}esponsiveness \textbf{t}uning for low-rank adaptation (\textbf{Heart-LoRA}). The proposed method explores redundancy among the heads and selectively activates task-responsive heads, thus enabling fine-grained head-level tuning. Additionally, given the different responsiveness of heads to diverse visual tasks, our proposed method dynamically activates a subset of the approximated heads that are tailored to the current task. Experimental results show that Heart-LoRA yields superior performance over state-of-the-art PETL approaches on visual adaptation benchmark datasets.
\end{abstract}

%%%%%%%%% BODY TEXT

\section{Introduction}
\label{intro}
Many works are currently focusing on adapting large models like Vision Transformer (ViT) \cite{dosovitskiy2020ViT}, pre-trained on vast datasets such as ImageNet \cite{deng2009imagenet}, to various tasks. However, the high parameter count in these models leads to substantial computational costs and often sub-optimal performance, limiting their applicability across diverse applications.

To address these issues, numerous studies are focusing on \textit{parameter efficient transfer learning} (PETL) methods like \cite{houlsby2019parameter} \cite{rebuffi2017learning} \cite{luo2023RepAdapter}. These methods efficiently adapt large pre-trained models by tuning a subset of parameters, outperforming full fine-tuning on various tasks with fewer parameters and lower computational demands. For instance, adapter-based PETL methods involve inserting trainable lightweight adapters into the model and keeping the model frozen as the backbone \cite{hu2021lora} \cite{jie2022convolutional}. These adapters are basically \textit{fully connected layers} with small hidden dimensions. During the adaption process, only these adapters need to be trained, while the rest of the parameters remain unchanged, thus reducing computational and storage costs.

\begin{figure}[t]
    \centering
    \includegraphics[width=0.45\textwidth]{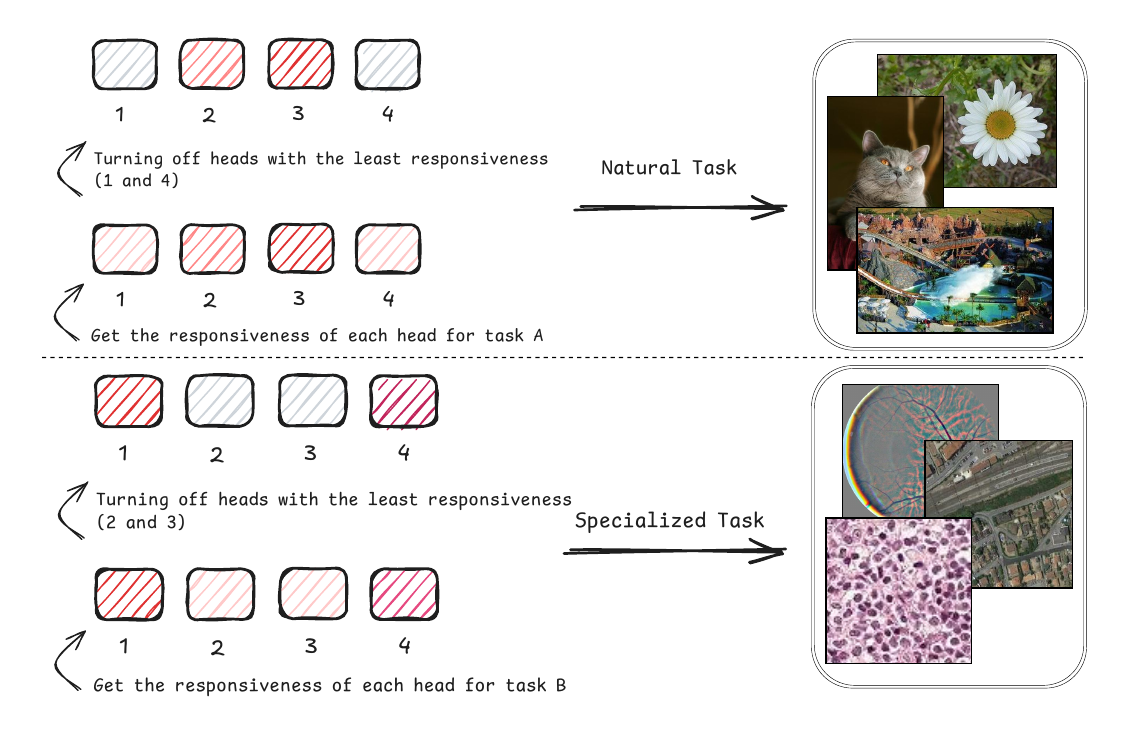} % 包含图片
    \caption{An example of recognising the responsiveness of the heads and then toggling the set of active heads across diverse tasks with different characteristics. Here we choose the example image for natural task and specialized task from VTAB-1K benchmark. The example for natural images assumes head 1 and 4 have the 2 least (since $ne=2$ here) responsiveness, while for specialized images those are head 2 and 3.  Once recognized, these heads are switched off during the adaptation process to eliminate their influence. Therefore the set of active heads are $\{2, 3\}$ and $\{1,4\}$, respectively.} % 图片标题
    \label{fig:multi_plots} % 图片标签
\end{figure}

Low-Rank Adaptation (LoRA \cite{hu2021lora}), as one of the most representative PETL methods, typically adds low-rank adapters to the transformation modules such as q and v in the multi-head self-attention (MHSA) module of transformers. For a specific module like q, the product of a pair of adapters can approximate the weight update for it. However, due to the design of MHSA, these modules include multiple heads (12 in ViT), and thus the weight updates approximated by the adapters also encompass updates for these multiple heads.

However, is the pre-configured multi-head setup in Transformer necessarily essential? For the Transformer architecture itself, various researches have examined the redundancy in attention heads through attention patterns \cite{kovaleva2019revealing}, pruning attention heads \cite{voita2019analyzing} \cite{clark2019does} \cite{michel2019sixteen_head_pruning_1}, attention visualization \cite{vig2019analyzingTheStructureOfAttentionInTransformer}, and probing tests \cite{clark2019does}. It has been proven that many attention heads produce similar attention matrices, exhibiting similar behaviors, and there is a limited set of attention patterns, indicating overall over-parameterization in MHSA \cite{clark2019does} \cite{kovaleva2019revealing}. These findings suggest that not all heads in the MHSA are essential, and this observation extends to the approximated heads in LoRA as well. Despite the parameter-efficiency of PETL techniques like LoRA, this redundancy still incurs problems like unnecessary storage and computational overhead. Firstly, we are required to store those less significant heads, which incurs additional storage costs. Secondly, during computation, data still pass through these heads, which do not significantly impact the model's performance. This leads to an increase in unnecessary computational costs as well.

Therefore, for the task currently being processed, deactivating those redundant heads could tackle the problem. We define a head's \textit{responsiveness} to the task as the contribution and activity level of the head to the current task. For heads with low responsiveness we selectively deactivate them during adaptation. Due to the diverse nature of tasks (for example, in VTAB-1K, tasks classified as Natural deal with images captured by cameras from natural scenes, while Specialized tasks deal with images captured using specialist equipment, such as medical images), the responsiveness of the same head can vary across diverse tasks. Therefore, by identifying the task-specific responsiveness of heads and accordingly activating or deactivating them, we can achieve a more comprehensive and profound understanding of the tasks.

In this paper, we propose \textbf{Hea}d-level \textbf{r}esponsiveness \textbf{t}uning in low-rank adaptation (\textbf{Heart-LoRA}). For a given task, the process involves obtaining the task-specific responsiveness of heads. We then selectively toggle the activation of heads based on their demonstrated effectiveness for the specific task at hand. This targeted approach ensures that only the most task-relevant heads are engaged, optimizing the model's performance and resource utilization. Our contributions are:

\begin{itemize}
    \item We propose Heart-LoRA, a novel task-aware PETL method which explores the \textit{head redundancy} in vision transformers. The proposed method enhances low rank adaptation with fine-grained, head-level responsiveness tuning across tasks.
    \item An efficient approximation strategy is introduced for calculating the responsiveness of different heads to given tasks, which recognizes the heads that have minimal impact on the task’s performance,.
    \item Extensive experiments on visual task adaptation benchmarks demonstrate that the proposed method achieves superior performance over state-of-the-art PETL methods in terms of both parameter efficiency and averaged accuracy.
\end{itemize}

\section{Related Work}

\subsection{Parameter-Efficient Transfer Learning}
Parameter-Efficient Transfer Learning (PETL) methods focus on adapting large pre-trained models like Vision Transformer (ViT) \cite{dosovitskiy2020ViT} to various downstream tasks by tuning only a small number of parameters in the model, thus achieving both parameter and storage efficiency. Prior work can be roughly divided into methods that involve inserting tokens and adapters. Token-based methods include \cite{jia2022VPT} \cite{lu2022promptDistributionLearning} \cite{zhou2022learningToPromptForVision-LanguageModeels_prompt_based_method5} \cite{shen2024multitask_prompt_based_method3} \cite{zang2022unified_prompt_based_method4} \cite{zhou2022conditional_prompt_based_method_6}, which essentially insert trainable tokens into the tokens in Multi-Head Self Attention (MHSA) block. Adapter-related methods, on the other hand, insert lightweight adapters into the model, either in the MHSA or in the Feed-Forward Network (FFN) block \cite{hu2021lora} \cite{chen2022adaptformer} \cite{houlsby2019parameter} \cite{jie2023fact} \cite{jie2022convolutional} \cite{pfeiffer2020adapter-P}. In full fine-tuning, whenever adapting to a new downstream task, a completely new backbone is required because full fine-tuning directly updates parameters in the original model. However, in the majority of PETL methods, during training, only the inserted tokens or adapters are updated while the original model remains frozen, allowing the model to serve as the backbone for multiple downstream tasks without additional parameters. We review some proposed PETL methods:

\textbf{AdaptFormer} \cite{chen2022adaptformer} introduces two parallel adapters to the FFN block, consisting of fully connected layers with weights $W_{down}$ and $W_{up}$, where $h$ is a small hidden dimension significantly less than $d$. \textbf{FacT} \cite{jie2023fact} tensorizes and decomposes the weight increments of ViT into lightweight factors, represented by $\Delta W$ decomposed into $U$, $V$, and $\Sigma$. \textbf{VPT} \cite{jia2022VPT} inserts trainable tokens into the input space, with VPT-Shallow adding tokens only to the first layer and VPT-Deep at every layer. \textbf{NOAH} \cite{zhang2024NOAH} combines multiple PETL adapters as prompt modules, optimizing their combination through a neural architecture search algorithm without manual design. \textbf{FourierFT} \cite{gao2024fourierft} employs the discrete Fourier transform for adaptation, whereas \textbf{MoRA} \cite{jiang2024MoRA} utilizes high-rank adapters along with non-parameterized compression and decompression operations for adaptation. \textbf{LoRA} introduces two low-rank adapters into the MHSA, adapting the transformation module within. \textit{Bi-LoRA} \cite{jie2023revisiting} is a quantized version of LoRA using low-precision adapters.

Methods like LoRA involve using inserted adapters to approximate the entire target module, where the target module could be the transformation module for q, k, v in MHSA. Due to the design of MHSA, essentially, the weight update approximation by LoRA for these modules also includes updates for all the heads. 

\subsection{Patterns among Attention Heads}

But does updating all the heads, thus utilizing all of them during the adaptation process really help? Prior research has extensively examined the attention mechanism and the essential design aspects of multi-heads in transformers, pointing out the problem of overparametrization and redundancy in the MHSA. Studies focusing on attention patterns \cite{kovaleva2019revealing}, head pruning techniques \cite{voita2019analyzing} \cite{clark2019does} \cite{michel2019sixteen_head_pruning_1}, visualization of attention \cite{vig2019analyzingTheStructureOfAttentionInTransformer}, and probing tests \cite{clark2019does} have shown that a considerable number of attention heads generate similar attention matrices, thus displaying redundant behavior. This suggests overall excess in parameters of MHSA \cite{clark2019does} \cite{kovaleva2019revealing}. Additionally, in specific scenarios, deactivating certain heads actually results in notable performance improvements \cite{kovaleva2019revealing}. 

Overall, these studies highlight the importance of selecting truly critical heads to handle the current task, since continuing to use these redundant heads during the training process incurs unnecessary computational and storage costs. Therefore, for different tasks, we should selectively activate or deactivate certain heads based on the unique characteristics of the task, focusing on those that are most responsive to the task and are most likely to make significant contributions. Overall, research on head-level responsiveness across diverse tasks helps us achieve more granular control of information during adaptation.

\section{Methodology}

\label{method}
In the methodology section, we describe two key steps of Heart-LoRA: firstly, obtaining the responsiveness of heads for current task, and secondly selectively activating or deactivating heads based on the obtained responsiveness, allowing for the toggling of active heads across diverse tasks. This process is illustrated in Fig. \ref{fig:multi_plots}.

\subsection{Task-specific Responsiveness Recognition}

Given a task, we aim to leverage information from heads that are critical and most likely to contribute to the task's performance, which also implies that their deactivation would lead to a decrease in performance. Concurrently, we seek to avoid utilizing information from heads that have minimal impact on the task's performance, as they represent the redundancy inherent in the Multi-Head Self Attention (MHSA). We define such a property as the \textit{responsiveness} of a head to the task, where heads with high responsiveness tend to make significant contributions to the task, while heads with low responsiveness can be deactivated without noticeable impact. In the MHSA, the transformations for the query, key, value, and output are parameterized by matrices $W_q, W_k, W_v, W_o \in \mathbb{R}^{N\times C}$, respectively. These modules are further segmented into $N_h$ heads due to the multi-head design: $\{W_q^{(i)}\}^{N_h}_{i=1}, \{W_k^{(i)}\}^{N_h}_{i=1}, \{W_v^{(i)}\}^{N_h}_{i=1}, \{W_o^{(i)}\}^{N_h}_{i=1}$. The calculation of MHSA for an input $X$ can be expressed as follows with $i$ indicating $i$-th head.
\begin{equation}
\begin{aligned}
    \textit{MHSA}(X)=&\\\sum_{i=1}^{N_h}\textit{softmax}&\left(\frac{XW_q^{(i)}{W_k^{(i)}}^\intercal X^\intercal}{\sqrt{C}}\right)XW_v^{(i)}{W_o^{(i)}}^\intercal
\end{aligned}    
\end{equation}
In LoRA, approximating the update $\Delta H \in \mathbb{R}^{N \times C}$ to a multi-head target module $H \in \mathbb{R}^{N \times C}$ with an intial value of $H_0 \in \mathbb{R}^{N \times C}$ in MHSA, where $H$ could be $W_q, W_k, W_v, W_o$,  with two low-rank adapters $A\in\mathbb{R}^{N \times d}$ and $B \in \mathbb{R}^{d \times C}$ can be expressed as:
\begin{equation}
    H=H_0+s\cdot\Delta H = H_0 + s\cdot AB
\end{equation} 
where $s$ is a scaling factor and $d$ is a small hidden-dimension. To obtain the task-specific responsiveness for each head, we divide $H$ as a set of all heads with $\forall h \in H,h \in \mathbb{R}^{\frac{N}{N_h} \times C}$. We use $R$ to denote a set of responsiveness score $r_i \in R ,\space i \in \{1,2,...,N_h\}$. For a specific responsiveness $r_i$ for a head $h_i \in H$, we define it as the change in loss function $L$ when this head is deactivated:
\begin{equation}
r_i=  L(h_i) - L(h_i=0) 
\label{e1}
\end{equation}
where we denote the deactivation of head $h_i$ as $h_i=0$. We can further approximate $L(h_i=0)$ using Taylor expansion, which given a function $f(x)$ at point $p$ is originally calculated as follows:
\begin{equation}
f(x)=\sum_{i=0}^{i}\frac{f^{(i)}(p)}{i!}(x-p)^{i}+R_i(x)
\end{equation}

Although higher-order Taylor expansions offer better approximation in mathematics, the heavy computational cost hinders their practical use. Besides, prior research \cite{molchanov2016pruningForResourseEfficientInference} \cite{spt} has demonstrated that the first-order Taylor expansion already possesses strong approximation capabilities. Moreover, due to the gradient descent algorithm, the gradient is readily obtainable during the backward pass. Therefore, we opt to use the first-order Taylor expansion to compute $r_i$ as:
\begin{equation}
    r_i  = \frac{\partial L}{\partial h_i}h_i -  \frac{\partial^2L}{\partial(h_i^2=\varepsilon)} \frac{h_i^2}{2}
    \label{eq core}
\end{equation}
The remainder term $\frac{\partial^2L}{\partial(h_i^2=\varepsilon)} \frac{h_i^2}{2}$ where $\varepsilon$ is between $0$ and $h_i$. The remainder is typically disregarded due to its substantial computational expense, as noted in \cite{gaikwad2018pruningSqueezenet} and \cite{molchanov2016pruningForResourseEfficientInference}. Unlike prior work \cite{spt}, although mathematically we obtain $r_i$ as $\frac{\partial L}{\partial h_i}h_i$, in practical applications, we also consider its opposite and absolute value due to considerations of positivity and negativity of the logits. This enables a more comprehensive calculation and consideration of the responsiveness.

The method we propose for calculating responsiveness boasts the advantages of being fast, simple, and intuitive. Unlike prior work \cite{spt}, we assess the responsiveness of heads approximated by adapters without resorting to approximations, instead we directly utilize the actual weights and gradients from the adapters. This allows us to have a direct grasp of the tuning conditions of the adapters. Overall, a single quick calculation per-task yield the task-specific responsiveness of all heads, laying the foundation for identifying low-responsive heads and subsequently deactivating them.

\subsection{Head-level Responsiveness-driven Deactivation}

After obtaining responsiveness $R$, we gain a comprehensive understanding of the activity level and contribution of all heads in the current task. Based on this understanding, we can selectively deactivate the heads. We refer to the distribution of active and inactive heads for certain tasks as the model's head-responsiveness pattern $P$. If a head $h_i\in H$ is active, we represent its corresponding element $p_i \in P$ with number 1, while number 0 indicating inactive head.
Based on $R$, we use $ne$ to indicate the number of heads that need to be deactivated. Therefore, the $ne$ heads with the smallest $r$ form a set denoted as $C$. For each head  $h_i \in C$, we deactivate it by multiplying it with corresponding $p_i \in P$:
\begin{equation}
  \forall i \in \{1,2,...,N_h\} \quad h_i  = h_i \cdot p_i
 \label{mask}
\end{equation}
When $h_i$ is in $C$, the corresponding $p_i$ is set to 0, thus the head is essentially masked. If $h_i$ is not in $C$, $p_i$ is 1 and does not cause any change to the behavior of the head. The implementation is in section F.1 of the appendix.

For the given task, we use a very short period as a warm-up to get the actual weight and gradient of the adapters, during which all heads are turned on (i.e., $\forall p \in P, \, p = 1$). This helps us gather comprehensive information about the behavior of all heads approximated by the adapters, rather than approximating the behaviour of adapters without actually adapting it \cite{spt}. After obtaining the responsiveness $R$, we perform the deactivation in Eq. \ref{mask}. Consequently, the results are contributed by the active heads in $P$.

For different tasks, the responsiveness $R$ varies, so does the head pattern $P$. This is an intuitive phenomenon: different tasks tend to process different types of objects. For example, VTAB-1K's Natural category focuses on standard camera images, whereas specialized tasks involve images from specific equipment like remote sensing. We also validate this premise in \ref{vary}. By adopting this granular approach to obtain specific responsiveness for each task, we fully consider the unique characteristics of different tasks, thereby helping us achieve better fine-tuning results across diverse tasks.

\begin{table*}[t]
\begin{center}
\setlength{\tabcolsep}{0.3pt}
\scalebox{0.85}{
\begin{tabular}{p{3cm}<{}p{1.25cm}<{\centering}p{0.75cm}<{\centering}|p{0.75cm}<{\centering}p{0.75cm}<{\centering}p{0.75cm}<{\centering}p{0.75cm}<{\centering}p{0.75cm}<{\centering}p{0.75cm}<{\centering}p{0.75cm}<{\centering}|p{0.75cm}<{\centering}p{0.75cm}<{\centering}p{0.75cm}<{\centering}p{0.75cm}<{\centering}|p{0.75cm}<{\centering}p{0.75cm}<{\centering}p{0.75cm}<{\centering}p{0.75cm}<{\centering}p{0.75cm}<{\centering}p{0.75cm}<{\centering}p{0.75cm}<{\centering}p{0.75cm}<{\centering}}
\toprule
\multicolumn{3}{c|}{}&\multicolumn{7}{c|}{\textbf{Natural}}&\multicolumn{4}{c|}{\textbf{Specialized}}&\multicolumn{8}{c}{\textbf{Structured}}\\
&\multicolumn{1}{c}{\STAB{\rotatebox[origin=c]{90}{Size (MB)}}}
&\multicolumn{1}{c|}{\STAB{\rotatebox[origin=c]{90}{Avg. Acc.}}}
&\multicolumn{1}{c}{\STAB{\rotatebox[origin=c]{90}{Cifar100}}}
&\multicolumn{1}{c}{\STAB{\rotatebox[origin=c]{90}{Caltech101}}}
&\multicolumn{1}{c}{\STAB{\rotatebox[origin=c]{90}{DTD}}}
&\multicolumn{1}{c}{\STAB{\rotatebox[origin=c]{90}{Flower102}}}
&\multicolumn{1}{c}{\STAB{\rotatebox[origin=c]{90}{Pets}}}
&\multicolumn{1}{c}{\STAB{\rotatebox[origin=c]{90}{SVHN}}}
&\multicolumn{1}{c|}{\STAB{\rotatebox[origin=c]{90}{Sun397}}}
&\multicolumn{1}{c}{\STAB{\rotatebox[origin=c]{90}{Camelyon}}}
&\multicolumn{1}{c}{\STAB{\rotatebox[origin=c]{90}{EuroSAT}}}
&\multicolumn{1}{c}{\STAB{\rotatebox[origin=c]{90}{Resisc45}}}
&\multicolumn{1}{c|}{\STAB{\rotatebox[origin=c]{90}{Retinopathy}}}
&\multicolumn{1}{c}{\STAB{\rotatebox[origin=c]{90}{Clevr-Count}}}
&\multicolumn{1}{c}{\STAB{\rotatebox[origin=c]{90}{Clevr-Dist}}}
&\multicolumn{1}{c}{\STAB{\rotatebox[origin=c]{90}{DMLab}}}
&\multicolumn{1}{c}{\STAB{\rotatebox[origin=c]{90}{KITTI-Dist}}}
&\multicolumn{1}{c}{\STAB{\rotatebox[origin=c]{90}{dSpr-Loc}}}
&\multicolumn{1}{c}{\STAB{\rotatebox[origin=c]{90}{dSpr-Ori}}}
&\multicolumn{1}{c}{\STAB{\rotatebox[origin=c]{90}{sNORB-Azim}}}
&\multicolumn{1}{c}{\STAB{\rotatebox[origin=c]{90}{sNORB-Ele}}}\\
\specialrule{0em}{1pt}{1pt}
\hline
\specialrule{0em}{1pt}{1pt}
\multicolumn{22}{l}{\emph{Conventional Fine-Tuning}}\\
\hline
\specialrule{0em}{1pt}{1pt}
\sc Full&327&68.9&68.9&87.7&64.3&97.2&86.9&87.4&38.8&79.7&95.7&84.2&73.9&56.3&58.6&41.7&65.5&57.5&46.7&25.7&29.1 \\
\sc Linear&0&57.6&64.4&85.0&63.2&97.0&86.3&36.6&51.0&78.5&87.5&68.5&74.0&34.3&30.6&33.2&55.4&12.5&20.0&9.6&19.2\\
%\specialrule{0em}{1pt}{1pt}
\hline
\specialrule{0em}{1pt}{1pt}
\multicolumn{22}{l}{\emph{PET methods}}\\
\hline
\specialrule{0em}{1pt}{1pt}
{\sc VPT-Deep}&2.03&72.0&\bf78.8&90.8&65.8&98.0&88.3&78.1&49.6&81.8&\bf96.1&83.4&68.4&68.5&60.0&46.5&72.8&73.6&47.9&32.9&37.8 \\
\text{\color{gray}{\sc NOAH}$^\dag$}&\text{\color{gray}1.37}&\text{\color{gray}75.5}&\text{\color{gray}69.6}&\text{\color{gray}\bf92.7}&\text{\color{gray}70.2}&\text{\color{gray}99.1}&\text{\color{gray}90.4}&\text{\color{gray}86.1}&\text{\color{gray}53.7}&\text{\color{gray}84.4}&\text{\color{gray}95.4}&\text{\color{gray}83.9}&\text{\color{gray}\bf75.8}&\text{\color{gray}82.8}&\text{\color{gray}\bf68.9}&\text{\color{gray}49.9}&\text{\color{gray}\bf81.7}&\text{\color{gray}81.8}&\text{\color{gray}48.3}&\text{\color{gray}32.8}&\text{\color{gray}44.2}\\
{\sc FourierFT}&1.20&72.2&65.3&89.9&69.7&99.0&90.8&81.5&54.8&84.4&93.5&80.6&74.5&73.3&59.0&43.8&77.2&73.7&49.3&26.4&34.0\\
{\sc MoRA}&1.15&75.1&72.6&89.8&72.0&99.1&91.0&89.1&55.7&86.6&94.0&84.7&74.3&78.3&62.8&49.7&78.6&82.2&51.3&34.5&35.1\\
{\sc LoRA}&1.13&76.4&72.0&91.2&71.6&99.1&91.3&88.9&56.4&87.2&94.6&83.9&74.9&\bf83.7&64.0&52.3&81.2&84.8&53.3&\bf38.1&43.4\\
{\sc SSF}&0.78&75.7&69.0&92.6&\bf75.1&\bf99.4&\bf91.8&\bf90.2&52.9&87.4&95.9&\bf87.4&75.5&75.9&62.3&\bf53.3&80.6&77.3&54.9&29.5&37.9\\
{\sc Adapter-P}&0.56&75.5&73.2&90.1&69.6&99.2&91.1&84.9&56.0&86.6&94.8&82.5&\bf75.8&82.9&63.9&49.7&79.7&81.7&55.5&31.6&42.2 \\
{\sc AdaptFormer}&0.56&\bf76.7&73.8&92.3&72.7&99.3&91.6&89.1&56.5&\bf87.8&95.5&84.9&75.2&83.3&62.5&52.4&\bf81.7&86.2&\bf55.9&34.4&40.2\\
{\sc BitFit}&0.39&65.2&72.8&87.0&59.2&97.5&85.3&59.9&51.4&78.7&91.6&72.9&69.8&61.5&55.6&32.4&55.9&66.6&40.0&15.7&25.1\\
{\sc FacT-TT}&0.30&\bf76.7&73.4&91.0&72.4&99.2&91.4&90.1&\bf56.6&87.3&94.7&84.5&\bf75.8&83.0&64.9&51.3&81.4&\bf87.4&53.2&33.5&\bf44.3\\
{\sc VPT-Shallow}&0.24&67.8&77.7&86.9&62.6&97.5&87.3&74.5&51.2&78.2&92.0&75.6&72.9&50.5&58.6&40.5&67.1&68.7&36.1&20.2&34.1\\
{\sc Compacter}&0.15&74.2&71.9&89.0&69.7&99.1&90.7&82.7&56.1&86.0&93.5&82.4&75.3&80.2&63.4&47.4&77.2&78.1&53.5&27.3&39.8\\
{\sc Bi-LoRA}&\bf0.14&\bf76.7&72.1&91.7&71.2&99.1&91.4&\bf90.2&55.8&87.0&95.4&85.5&75.5&83.1&64.1&52.2&81.3&86.4&53.5&36.7&44.4\\
%{\sc FacT-TT}*~\cite{fact}&0.14&75.3&71.3&89.6&70.7&98.9&91.0&87.8&54.6&85.2&95.5&83.4&75.7&82.0&\bf69.0&49.8&80.0&79.2&48.4&34.2&41.4\\
%{\sc FacT-TT}~\cite{fact}&\bf0.14&76.2&73.2&89.6&72.4&99.1&91.2&88.4&\bf56.6&87.0&94.6&83.3&\bf75.8&82.2&64.7&50.9&81.3&86.1&52.5&33.1&44.1\\
%SparseAdapter~\cite{sparseadapter}\\
\hline
\specialrule{0em}{1pt}{1pt}
\rowcolor{gg} {\sc Heart-LoRA}&\bf0.11&\bf77.2&72.7&\bf92.3&71.4&99.2&91.4&\bf90.2&55.9&\bf88.0&\bf95.8&85.6&75.5&\bf83.7&64.9&52.3&\bf82.3&\bf86.7&53.5&\bf40.0&\bf44.8\\
\bottomrule
\end{tabular}
}
\end{center}
\caption{\textbf{Results on the VTAB-1K benchmark}. "Avg. Acc." signifies the mean performance across three categories. "Size" indicates the average number of trainable parameters within the backbones for each task, excluding the classification heads which account for 0.14 MB per task on average. $^\dag$ indicates results sourced from~\cite{zhang2024NOAH} where inputs are standardized.}
\label{vtab}
\vspace{0pt}
\end{table*}

\section{Experiments}

We denote our method, which includes first recognizing the responsiveness pattern of the heads and then deactivating those with the least responsiveness, as \textbf{Hea}d-level \textbf{r}esponsiveness \textbf{t}uning for low-rank adaptation (\textbf{Heart-LoRA}). First, we discuss some implementation details of Heart-LoRA, and then evaluate it using ViT on the VTAB-1K benchmark. We then evaluate it in the context of few-shot learning. Following that, we also evaluate it on the Swin Transformer architecture using VTAB-1K, further illustrating the architecture-agnostic nature of the method. Finally, we analyze some variations of the method and the head patterns obtained from different tasks, offering some insights into how head-level responsiveness works.

\subsection{Adaptation on VTAB-1K Benchmark}

\subsubsection{Implementation Details}

Models like the Transformers comprises numerous layers, each equipped with its own Multi-Head Self-Attention (MHSA) module. For the sake of simplicity, the approach we adopt is to accumulate the responsiveness of all the heads across all the layers. This means that we repeat $N_l$ times, each time in a layer, to obtain final $R$, where $N_l$ represents the number of layers in the model. After obtaining $C$, we apply the same deactivation strategy to each layer as well. This implementation can fully consider the characteristics of different layers, thereby providing a more comprehensive understanding. We also utilize the quantization technique proposed in \cite{jie2023revisiting} to ensure the storage efficiency of the implementation. Further information can be found in the appendix.

\subsubsection{Datasets} We first evaluate our method on the VTAB-1K benchmark \cite{zhai2019vtab}, a suite of vision tasks designed to evaluate general visual representations, thus demonstrating our method's general capability. The VTAB-1K consists of 19 tasks from various domains and with various semantics, divided into three groups: 1 ) \textbf{Natural} 2) \textbf{Specialized} 3) \textbf{Structured}. For each task, VTAB-1K uses only 1000 examples. All our reported results are top-1 accuracy on the test sets. Following \cite{jie2023revisiting} \cite{jie2023fact} \cite{zhang2024NOAH} \cite{jia2022VPT}, we use unnormalized input for all datasets. Some previous work which normalizes input is re-implemented by \cite{jie2023revisiting}, so in such case we directly refer to their original data.

\subsubsection{Comparison with Current PETL Methods} we compare our method with \textbf{VPT} \cite{jia2022VPT}, \textbf{NOAH} \cite{zhang2024NOAH}, \textbf{AdaptFormer} \cite{chen2022adaptformer}, \textbf{BiTFit} \cite{zaken2021bitfit}, \textbf{FacT-TT} \cite{jie2023fact} \textbf{FourierFT} \cite{gao2024fourierft}, \textbf{MoRA} \cite{jiang2024MoRA} and \textbf{LoRA} \cite{hu2021lora} and \textbf{Bi-LoRA} \cite{jie2023revisiting}. The hidden dimension $r$ is set to 8 for AdaptFormer and LoRA. FacT-TT's rank $r$ is searched from $\{8,16,32\}$. We implement both FourierFT and MoRA, ensuring that they have the same order of parameter count and identical training hyper-parameters as LoRA. Other methods' settings follow their best recipes in papers. For Heart-LoRA, the hyper-parameter $ne$ is roughly searched from $\{1,3,6\}$ empirically. Following \cite{jie2023revisiting} and \cite{zhang2024NOAH}, we use AdamW as optimizer with a learning rate of 1e-3 and a batchsize of 64 to train for 100 epochs. We first train for 10 epochs as warm up to get the responsiveness and set the $C$ for the remaining 90 epochs. For backbone, following \cite{jie2023revisiting} \cite{jie2023fact}, all methods use ViT-B/16 pre-trained on Image-Net 21K.

\begin{figure*}[t]
    \centering
    \includegraphics[width=1\textwidth]{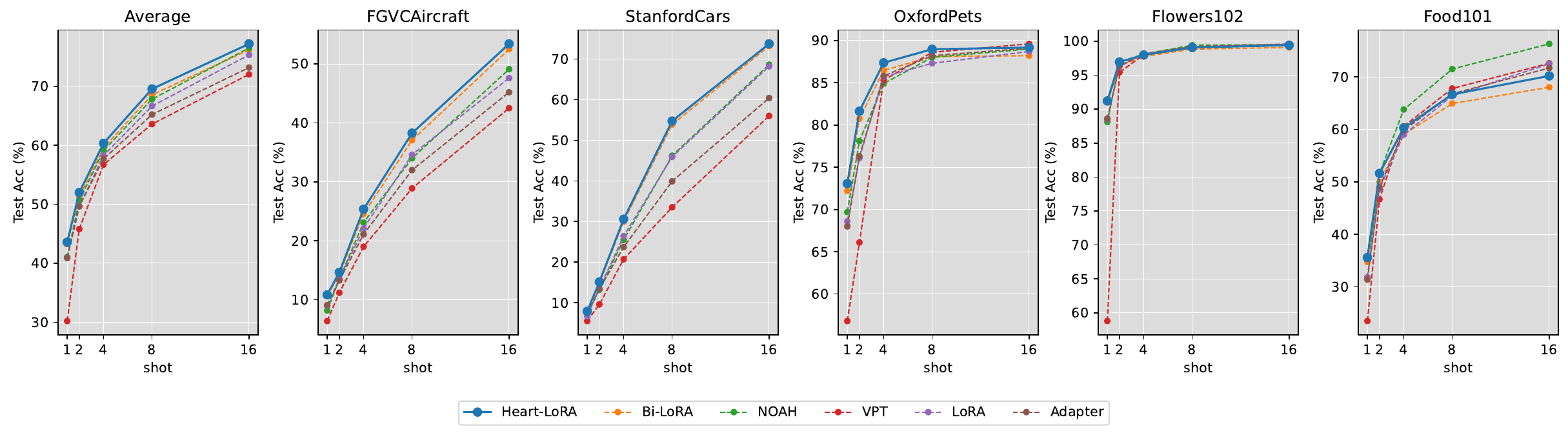} % 包含图片
    \caption{Full results on few-shot learning. Compared methods include NOAH, Bi-LoRA, VPT, LoRA, and Adapter. All experiments are conducted under a setting with shots ranging from 1 to 16. The results for each shot are averages over three distinct seeds.} % 图片标题
    \vspace{0pt}
    \label{few-shot}
\end{figure*}

\subsubsection{Results} All our results are shown in Table \ref{vtab}, in which Heart-LoRA outperform all existing PETL methods on VTAB-1K benchmark. It outperform Bi-LoRA by $0.5\%$ while utilizing fewer parameters. Performance of Heart-LoRA also exceeds LoRA by $0.8\%$, which can be regarded as the baseline for the PETL method on MHSA, which is exactly the focus of our study. To be specific, Heart-LoRA achieve state-of-the-art performance on 8 tasks across 19 datasets. from which we can conclude:

(1) Reducing the redundancy among attention heads in MHSA by deactivating heads with the least responsiveness can bring performance improvement on various domains. Therefore, any other methods utilizing the MHSA block in transformer could potentially benefit from our method based on an empirical inference: our method can improve performance by reducing redundancy among attention heads. Therefore, it is possible for our method to be combined with other PETL methods that focus on reducing different types of redundancy in PETL, thereby achieving even higher performance, since the effect of reducing redundancy on performance could be additive.

(2) Heart-LoRA matches LoRA and Bi-LoRA in terms of parameter efficiency by not adding any extra structures or parameters, avoiding the need for solely parameter-increasing methods to enhance accuracy. A core mechanism of Heart-LoRA is its selective processing of information at the head-level directly on the input, a process that doesn't influence the model structure. Instead, it obtain higher performance by rather discarding unimportant information.

(3) Although PETL adapters are in use, the storage and computation costs brought about by head redundancy still prevent us from fully realizing the potential of PETL. Heart-LoRA addresses these issues through selective deactivation: since there is no need to store the identified redundant heads, for each task, we can save $\frac{ne}{N_h}$ parameters (0.11MB versus 0.14MB). Additionally, as data does not need to pass through deactivated heads, in the case of $ne=6$, we can save half of the computation cost, which helps to accelerate the adaptation process.

Through extensive experiments, we find that by deactivating a small number of heads such as $ne=1/3$ , we can often achieve the same accuracy without the deactivation process and in many cases we can improve the performance. Most of the best results come from $ne=1$ or $ne=3$, which makes sense because having too many deactivated heads can also lead to the loss of essential information for the task, rather than just reducing the redundancy. Previous work has proven that many heads still play an important and consistent role, pruning to many of them out will have a detrimental effect on the overall performance \cite{kovaleva2019revealing} \cite{voita2019analyzing}. The upper limit $ne=6$ of our search space is determined through a simple criterion of deactivating at most half of the heads (ViT has 12 heads in the MHSA), and exceeding this upper-bound could potentially results in noticeable performance degradation in certain tasks since more than half of the information from the heads is discarded. \textbf{Further analysis are provided in the subsequent section \ref{heads} and section C in appendix}. In general, we observe that Heart-LoRA successfully recognise problems of head redundancy even in PETL adapters like LoRA: extra storage cost and computation cost with potential inferior performance, and tackle these challenges accordingly with higher performance and competitive parameter-efficiency. 

\subsection{Few Shot Learning} Few-shot learning evaluates a model's proficiency in learning from minimal training data and generalizing to new tasks, a critical capability in scenarios where data acquisition is challenging. We conducted experiments on five datasets: FGVC-Aircraft \cite{maji2013aircraft}, Oxford-Pets \cite{parkhi2012Pets}, Food-101 \cite{bossard2014food}, Stanford Cars \cite{krause2013cars}, and Oxford-Flowers102 \cite{nilsback2006flowers}, using various shot settings (1, 2, 4, 8, and 16 shots) with three distinct seeds. In comparing Heart-LoRA with current PETL methods such as Bi-LoRA, LoRA, VPT, NOAH, and AdaptFormer, we set the hyper-parameter \( h \) for LoRA and AdaptFormer to 8. For Heart-LoRA, results were based on a rough selection using either positive or negative values from the formula in Eq. \ref{eq core}, and we selected the top accuracy from \( ne=1 \) or \( ne=3 \) to showcase its performance in limited settings. Average results were computed from three distinct seeds for each shot setting.

As shown above in Figure \ref{few-shot}, Heart-LoRA outperform all of compared PETL methods, and achieve the state-of-the-art performance in average. Heart-LoRA performs the best on FGVC-Aircarft, Oxford-Pets, Stanford Cars and Oxford-Flowers102. On Food-101, Heart-LoRA shows slighly inferior performance compared to some methods. On average though, Heart-LoRA outperform Bi-LoRA and LoRA by a significant margin. Heart-LoRA surpass Bi-LoRA's performance by roughly $0.8\%$ in average, demonstrating that Heart-LoRA can still achieve high performance with almost no searching of the hyper-parameters even when only a limited number of training samples are available. 

\textbf{Additionally, more results on VTAB using different strategies and model architectures are provided in section A and section B in appendix}.

\begin{figure}[t]
    \centering
    \begin{minipage}{0.25\textwidth}
        \centering
        \includegraphics[width=\textwidth]{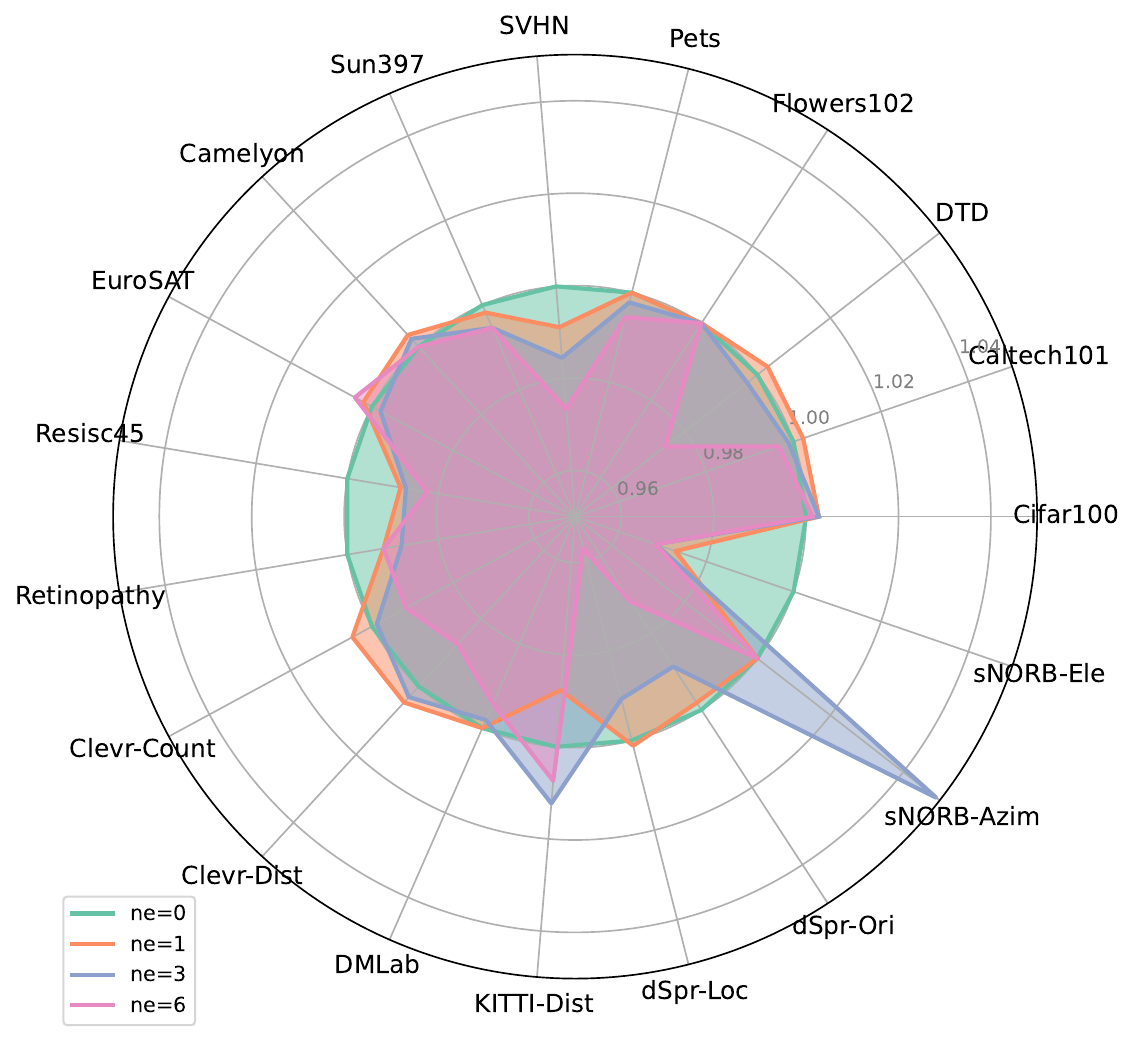} % 包含图片
        \caption{To test the effect of arbitrary deactivation, $ne$ of 1, 3, and 6 were applied on Heart-LoRA, and comparisons were made with the baseline without any deactivation. $ne=x$ means that in the Heart-LoRA, there are $x$ heads at the front being deactivated while $ne=0$ serves as the baseline. Values in the figure represent the ratio of results after applying deactivation to the baseline results.} % 图片标题
        \label{radar} % 图片标签
    \end{minipage}
    \hfill
    \begin{minipage}{0.2\textwidth}
        \centering
        \includegraphics[width=\textwidth]{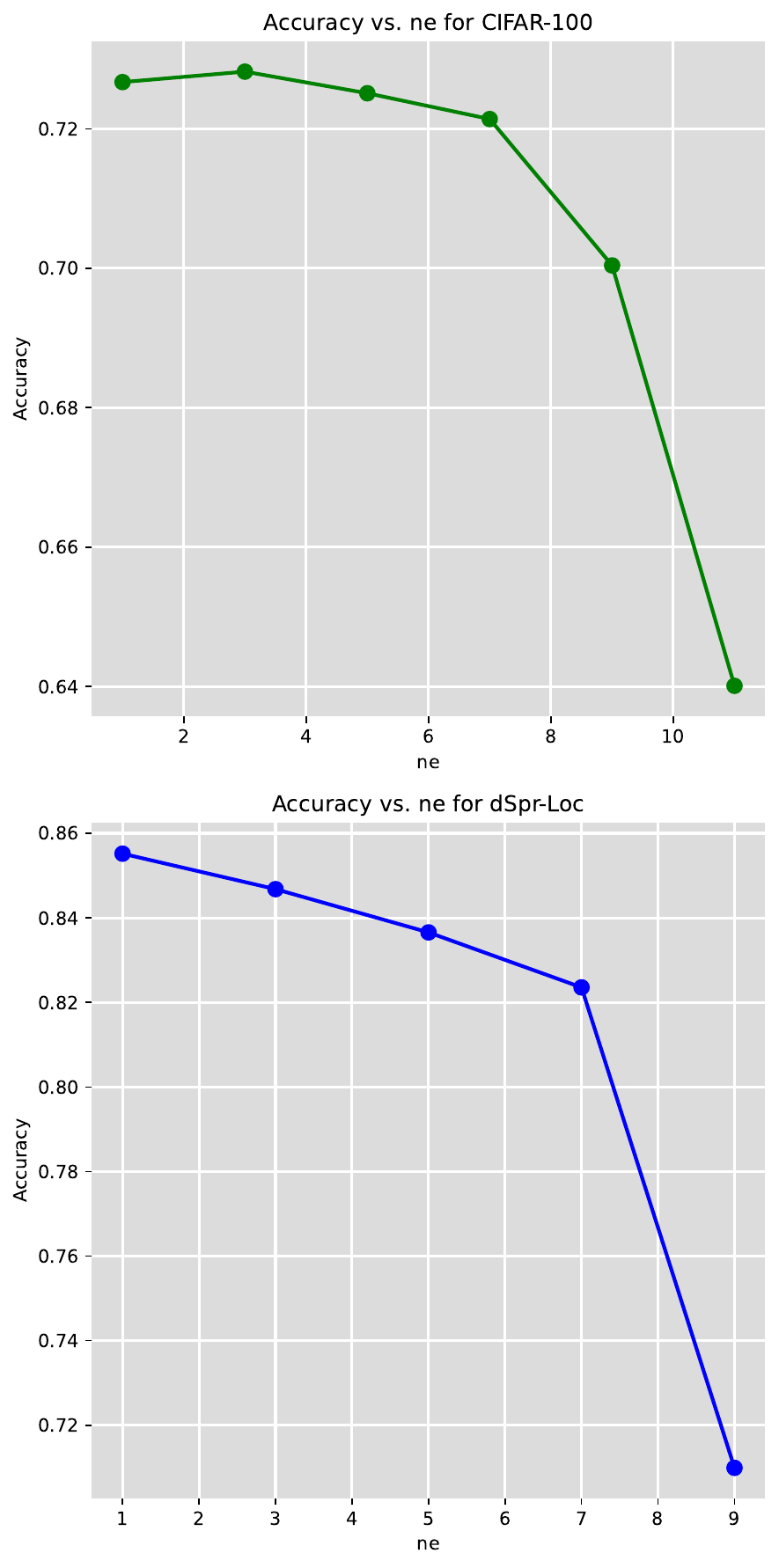} % 包含图片
        \caption{Impact of $ne$ on performance on CIFAR-100 and dSpr-Loc datasets.} % 图片标题
        \label{ne} % 图片标签
    \end{minipage}
    \vspace{0pt}
\end{figure}

\begin{figure}[ht]
    \centering
    \includegraphics[width=0.99\linewidth]{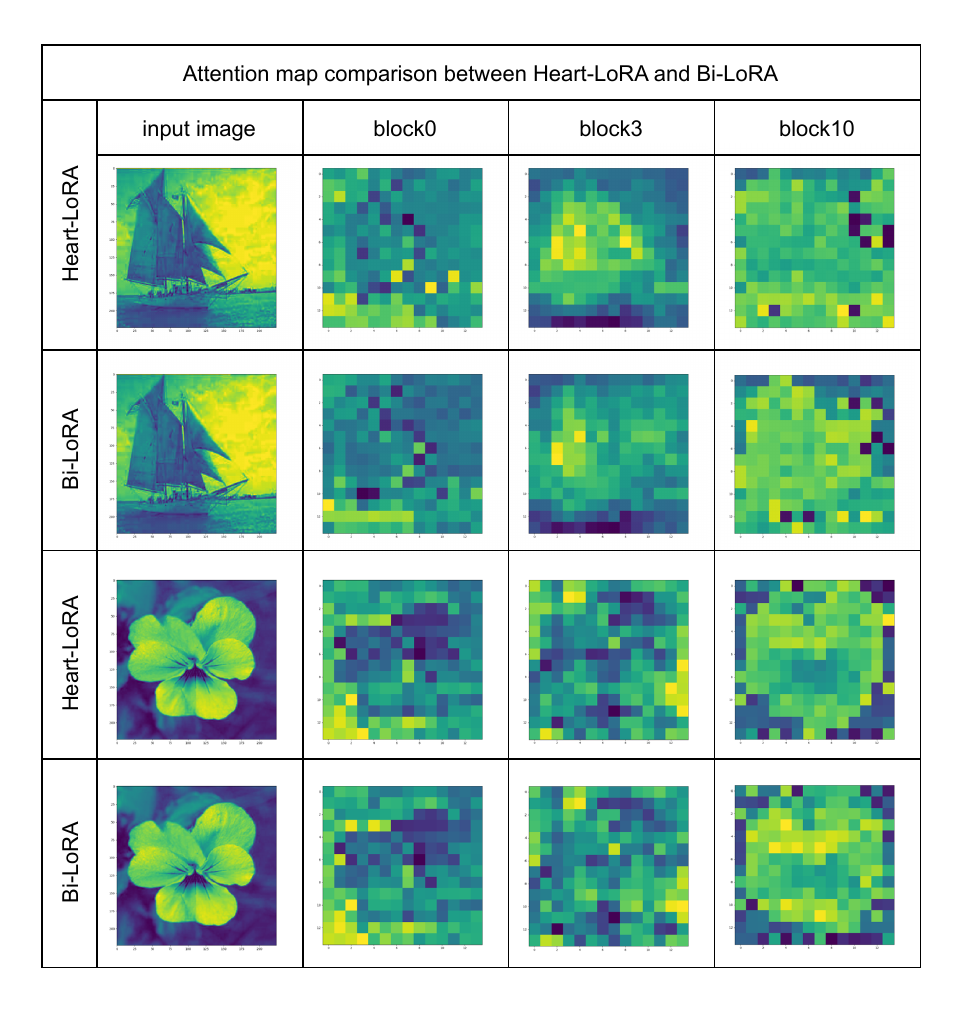}
    \caption{Comparison of attention map between Heart-LoRA and Bi-LoRA. For the example here the first image is taken from Caltech101 dataset while the second from Flowers102 dataset. We use $ne=3$ and $ne=6$, respectively for Hear-LoRA. It can be observed that the attention maps are nearly the same after certain deactivation, demonstrating the existence of head redundancy.}
    \label{fig:attention}
\end{figure}

\subsection{Understanding Head-level Responsiveness}

\subsubsection{Head Redundancy during Adaptation}
\label{arbitrary deactivation}

What motivates Heart-LoRA is the redundancy among the attention heads and therefore the extra storage and computation burden it brings. Since prior work has indicated that heads in MHSA exhibit redundancy, we also conduct experiments to study this phenomenon in the use of adapters in PETL methods to demonstrate its existence. To prove that the same phenomenon of many attention heads exhibiting similar behaviours can also be observed in the adaptation of ViT, we first test the effect on the performance by arbitrarily deactivating some heads. Specifically, we directly deactivate the $ne$ head that is positioned at the front. For example, if we specify $ne=3$, then head $1,2,3$ will be deactivated. This setup is based on the observation that if the redundancy phenomenon is prevalent, then this method of not distinguishing heads for diverse tasks should also be effective for certain tasks. Moreover, in addition to quantitative analysis, to gain a more intuitive qualitative understanding of head redundancy in adaptation, we analyze the attention maps using Heart-LoRA and Bi-LoRA as shown in Fig. \ref{fig:attention}.

From the results of arbitrary mask, we can see that even when the same deactivation strategy is applied to all tasks, we can still achieve notable performance gains on many tasks compared to typical low-rank adaptation. At the same time, by analyzing the attention maps, we find that even when parts of the heads are deactivated ($25\%$ in Caltech101 and $50\%$ in Flowers102), the attention maps of Heart-LoRA are \textit{nearly identical} to those when all heads are activated. This strongly indicates that deactivating heads does not notably affect the behavior of attention, thereby illustrating the redundancy nature of these heads. Overall, through both quantitative and qualitative analysis, we have fully demonstrated that the phenomenon of head redundancy still exists in PETL methods such as LoRA, validating the motivation behind our approach.

A direct comparison between arbitrary deactivation and Heart-LoRA is further provided in section E of the appendix.

\begin{figure}[t]
    \centering
    \includegraphics[width=0.99\linewidth]{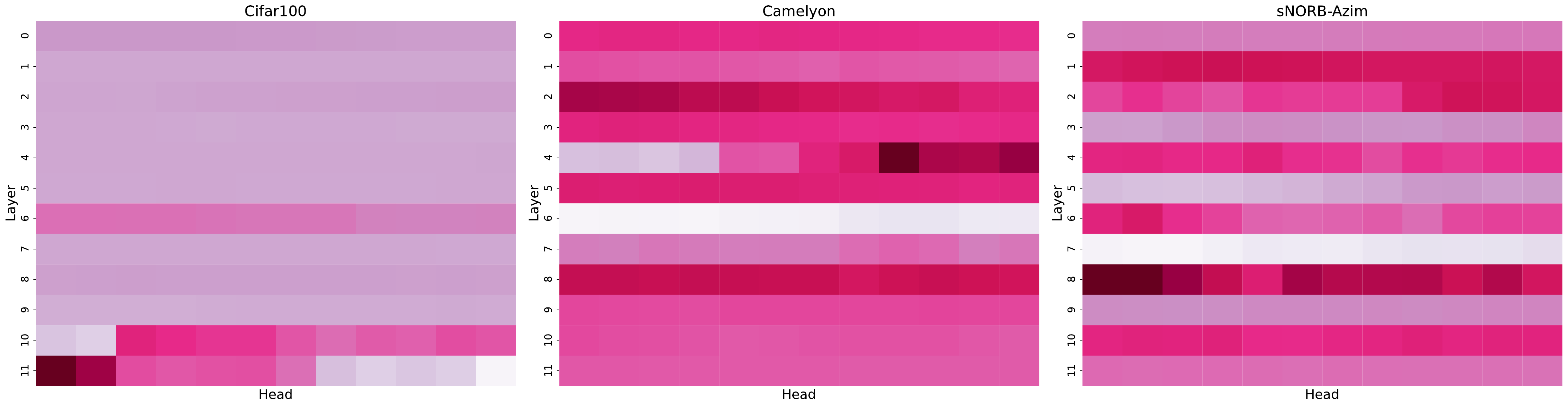}
    \caption{Head-level responsiveness pattern across 3 tasks of distinct category. It can be observed that the responsiveness pattern clearly varies dramatically across tasks, validating the premise of Heart-LoRA.}
    \label{fig:heatmap1}
\end{figure}

\begin{figure}[t]
    \centering
    \includegraphics[width=0.99\linewidth]{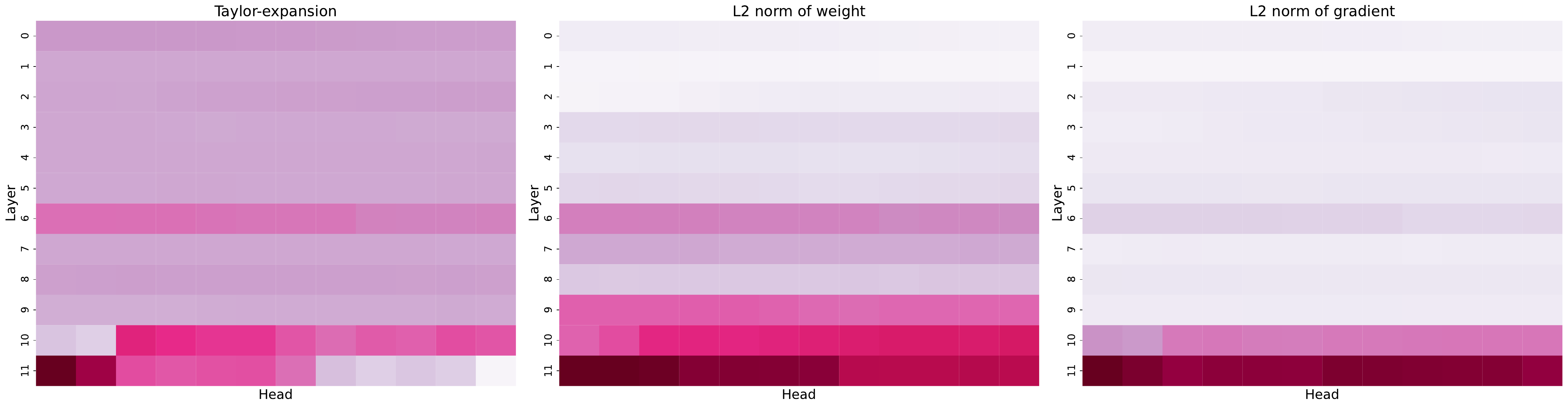}
    \caption{Comparison of obtained responsiveness using alternative methods of calculating responsiveness. Task here is Cifar100. The obtained responsiveness exhibits similar patterns.}
    \label{fig:heatmap2}
\end{figure}

\subsubsection{Head-level Responsiveness across Tasks}
\label{vary}
A distinctive feature of Heart-LoRA is its ability to identify \textit{task-specific} responsiveness across different tasks and subsequently perform targeted deactivation. This selectivity is based on the premise that responsiveness varies across tasks. Therefore, we selected one task from each of the categories: Natural, Specialized, and Structured, to cover a range of task types. For each task, we analyzed the responsiveness pattern of all heads in the query transformation matrices under adaptation. The visualization results are in Fig. \ref{fig:heatmap1}, where darker color indicates stronger responsiveness.

It is clear that we can observe significant differences in responsiveness patterns as the task types vary. On the Cifar100 dataset, the first half of the heads in the highest layer exhibit high responsiveness, while on Camelyon, it is the latter half of the heads in the middle layers, and on sNORB-Azim, it is the entire layer of index 8. These varied patterns validate the premise of Heart-LoRA that responsiveness varies across tasks.

\subsubsection{Responsiveness Pattern Using Alternative Approaches}

There are alternative methods to calculate responsiveness. We provide a detailed quantitative performance comparison in  section D of the appendix. Here we present a visualization of the responsiveness patterns in Fig. \ref{fig:heatmap2}.

We used the L2 norm of weight (referred to as weight) and the L2 norm of gradient (referred to as gradient) as alternative calculation methods for responsiveness. The results show that for the same task, the responsiveness patterns obtained using weight or gradient are similar to those of the Heart-LoRA, with highly responsive heads concentrated in the later layers. However, the typica Heart-LoRA uses Taylor expansion, which provides a better mathematical approximation and includes information from both weight and gradient, making the consideration more comprehensive. At the same time, the higher performance in section D of the appendix also indirectly demonstrates the superiority of our adopted approach.

\subsection{Ablation}
\begin{table}[t]
\centering
\resizebox{0.47\textwidth}{!}{
\begin{tabular}{l*{5}{c}}
\toprule
 & Natural & Specialized & Structured & Average \\\hline
 \specialrule{0em}{1pt}{1pt}
Heart-LoRA & 81.87 & 86.20 & 63.53 & 77.20 \\
\rowcolor{lightgray}Heart-LoRA (FP32) & 81.71 & 85.58 & 62.84 & 76.71 \textcolor{red}{($\downarrow 0.5$)} \\ \bottomrule
\end{tabular}
}
\vspace{0pt}
\caption{Comparison between the original Heart-LoRA using quantization and Heart-LoRA using FP32 precision adapters on VTAB-1K.}
\label{FP32}
\vspace{0pt}
\end{table}

\textbf{Quantization}: For the adapters used in the model structure of Heart-LoRA, we employed a quantization method to reduce computational and storage costs. Here, we analyze the impact of using FP32 precision adapters (i.e., without quantization) on the results of Heart-LoRA. From the Table \ref{FP32}, it can be observed that using FP32 precision adapters results in slight degradation in Heart-LoRA's performance, but it remains comparable to some competitive methods such as FacT and AdaptFormer. These results suggest that quantization plays a significant role in the Heart-LoRA approach, and utilizing quantization can help to better capitalize on the benefits of the Heart-LoRA method and attain improved performance.

\textbf{Number of Deactivated Heads}:  \label{heads} The effects of $ne$ on the performance are shown in Fig \ref{ne}. For the hyper-parameter $ne$, we study the results when $ne \in \{1,3,5,7,9,11\}$. When $ne$ exceeds half of the heads (In ViT-B it is 6), the performance generally experiences noticeable degradation. This is most likely due to the loss of essential information. When $ne \leq 6$, we mostly deactivate heads with redundant information, while $ne=6$ or above, we could potentially eliminate heads that are in fact not redundant or even crucial for a specific task. Also note that the performance will actually remains the same or even increases notably when a small fraction of heads are deactivated, as illustrated in Fig. 8 of the appendix. This is mostly due to the recognition and deactivation of redundant heads that contribute least to the current task, which is precisely the strength of Heart-LoRA.

Additionally, \textbf{we report full ablation results in section C in appendix}, providing a more comprehensive understanding of the impact of $ne$. 

\section{Conclusion}

In this paper, we investigate and explore the impact of task-specific head-level responsiveness within PETL methods such as low-rank adaptation (LoRA). We propose Heart-LoRA, a method that recognizes the responsiveness of heads according to the characteristics of different tasks, thus selectively enabling the activation or deactivation of heads to focus on important ones while ignoring redundant and unimportant heads. We validated the effectiveness of our approach on the VTAB-1K benchmark and in few-shot learning scenarios, and also conducted experiments on the Swin Transformer to demonstrate the architecture-agnostic nature of our method. We analyzed the performance and mechanisms of head-level responsiveness and conducted multiple experiments to support our findings. Our findings indicate that the studying attention-head patterns could provide valuable guidance for the design and application of PETL methods.

\clearpage

%%%%%%%%% REFERENCES
{\small
\bibliographystyle{ieee_fullname}
\bibliography{egbib}
}

\clearpage
\appendix
\clearpage

\section*{Appendix}

\setcounter{section}{0}

\begin{table}[b]
\centering
\resizebox{0.47\textwidth}{!}{
\begin{tabular}{l *{4}{c}} % 使用 c 列类型将数字居中对齐
\toprule
Method & Average & Natural & Specialized & Structured \\\hline
\specialrule{0em}{1pt}{1pt}

Full & 68.9 & 75.9 & 83.4 & 47.6 \\
LoRA & 76.4 & 81.5 & 85.2 & 62.6 \\
\rowcolor{gg} 
Heart-LoRA$^\dag$ & \textbf{76.8} \textcolor{ForestGreen}{($\uparrow 0.4$)} & \textbf{81.7} \textcolor{ForestGreen}{($\uparrow 0.2$)} & \textbf{85.8} \textcolor{ForestGreen}{($\uparrow 0.6$)} & \textbf{63.0} \textcolor{ForestGreen}{($\uparrow 0.4$)} \\ \bottomrule
\end{tabular}
}
\vspace{0pt}
\caption{Results on VTAB-1K of Heart-LoRA$^\dag$, using a layer by layer deployment of the calculation in section \ref{method}.}
\label{heartloraplus}
\end{table}

\section{Heart-LoRA with More Granularity}
\label{heart-lora-dag}
As mentioned earlier, the original Heart-LoRA uses layer-by-layer accumulation to calculate responsiveness. This approach ensures that the responsiveness of each head is considered globally. However, directly calculating responsiveness at each layer and applying deactivation is also worth exploring. Therefore, we propose a variant of Heart-LoRA that directly calculate responsiveness at each layer and then perform the deactivation at layer level. We term this variant as Heart-LoRA$^\dag$ and present its results in Table \ref{heartloraplus}. 

From the results, it can be seen that the layer-by-layer implementation of Heart-LoRA$^\dag$ also performs well, outperforming LoRA by $0.6\%$. We argue that this is due to our proposed responsiveness calculation algorithm, where the responsiveness of the heads acts on the local layer, which could potentially yield better results in certain tasks. Overall, both Heart-LoRA$^\dag$ and the original Heart-LoRA significantly outperform LoRA and generally require fewer parameters to function.

\begin{figure*}[t]
    \centering
    \includegraphics[width=\linewidth]{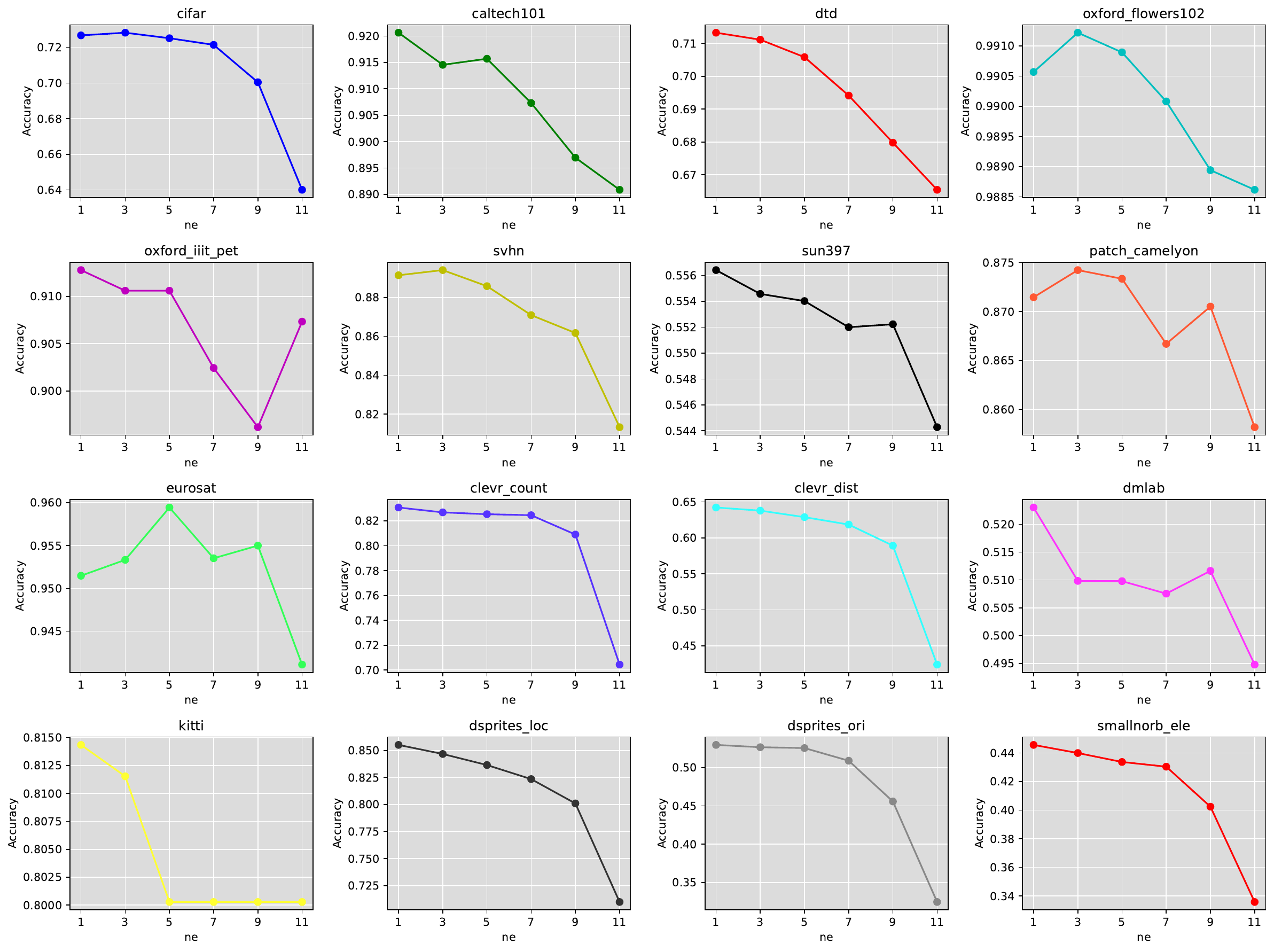}
    \caption{Full ablation results on the effect of $ne$ on different datasets in VTAB-1K benchmark.}
    \label{fig:full-ablation}
\end{figure*}

\begin{table}[b]
\centering
\resizebox{0.48\textwidth}{!}{
\begin{tabular}{l *{4}{c}} % 使用 c 列类型将数字居中对齐
\toprule
Method & Average & Natural & Specialized & Structured \\\hline
\specialrule{0em}{1pt}{1pt}

Full & 74.99 & 79.2 & 86.2 & 59.7 \\
Linear & 62.60 & 73.5 & 80.8 & 33.5 \\
VPT-Deep & 71.55 & 76.8 & 84.5 & 53.4 \\
Bi-LoRA & 76.67 & 82.0 & 86.8 & 61.2 \\
\rowcolor{gg} 
Heart-LoRA & \textbf{76.93} \textcolor{ForestGreen}{($\uparrow 0.26$)} & \textbf{82.3} \textcolor{ForestGreen}{($\uparrow 0.3$)} & \textbf{87.1} \textcolor{ForestGreen}{($\uparrow 0.3$)} & \textbf{61.4} \textcolor{ForestGreen}{($\uparrow 0.2$)} \\ \bottomrule
\end{tabular}
}
\vspace{0pt}
\caption{Results on VTAB-1K using Swin-B as backbone, reported accuracy from average, natural, specialized and structured.}
\label{table2}
\end{table}

\section{On Hierarchical Transformers}
\label{swin}
We also apply our method on Swin Transformer \cite{liu2021swintransformer}, a widely adopted and representative design of hierarchical structures of transformers aiming to improve ViT's computational efficiency.

Given that the number of heads and dimension of token is different across layers, simply setting a $ne$ hyper-parameter is not enough. Instead, we set a hyper-parameter $ratio$ indicating the percentage of heads to deactivate in the window attention. The accumulation of responsiveness $R$ and thus the candidate $C$ for swin transformer is restricted in layers with the same number of attention heads in this case, which makes sense since layers with the same number of attention heads generally are in the same hierarchy of Swin Transformer.

We use Swin-B as our backbone following \cite{jie2023revisiting}. It consists of four stages, with depths \{2, 2, 18, 2\} and number of heads \{4, 8, 16, 32\}. In this case, take $ratio=0.25$ as an example, we set the number of candidate of deactivation $C$ for the four stages separately as \{1, 2, 4, 8\}. While in many tasks, the deeper the layer is, it learns more task specific information, setting a fixed number of heads also proves to be effective since it discard little information in deeper layer while discarding much information in shallower layer where more redundancy persists. 

The results are shown in Table \ref{table2}, using \textbf{swin-B} pre-trained on ImageNet-21K \cite{deng2009imagenet} as backbone for Heart-LoRA. Following \cite{jie2023revisiting} \cite{jie2023fact}, we compare Heart-LoRA with several methods which could also be applied on swin transformer: Full, Linear, VPT \cite{jia2022VPT} and Bi-LoRA \cite{jie2023revisiting}. From the results, we observe that Heart-LoRA achieve high performance in Swin Transformer even with only very few hyper-parameter settings, showcasing its versatility across different transformer designs. While the performance may be sub-optimal compared to ViT, there is potential for further improvement as our experiments were conducted in only a limited number of cases. Additionally, our method on Swin-B also demonstrates that Heart-LoRA is structure-agnostic and method-agnostic, indicating its potential to enhance performance across various architectures.

\section{Full Ablation on Number of Heads}
\label{full-ablation}

We provide a comprehensive analysis of the impact of $ne$ settings on performance. We selected 16 datasets, which essentially encompass all types of tasks in the VTAB-1K benchmark. For each dataset, we varied the value of $ne$ and tracked the corresponding performance. Since we are using ViT-B, which has 12 heads, the values for $ne$ were set at $1, 3, 5, 7, 9, 11$ to provide more data points. The results are in Fig \ref{fig:full-ablation}.

The results show a general trend where increasing $ne$ leads to a decrease in performance. However, the degree of this decrease is not linear; in many datasets, we observed that the performance drop after increasing $ne$ is not significant (like in Cifar100 and dSpr-Ori), and in some cases the performance even improves (EuroSAT and SVHN). We argue that this phenomenon is due to the inherent redundancy among all the heads in the multi-head self-attention mechanism, where some heads have minimal responsiveness (defined in section \ref{intro}) and thus contribute little to the current task. Since Heart-LoRA can accurately identify these minimally responsive heads and deactivate them accordingly, the performance does not drop significantly, and in many datasets, it even improves.

However, when $ne$ is large, such as $ne=9$ or $ne=11$, more than half of the heads are deactivated. At this point, performance generally experiences significant degradation. This is because deactivating too many heads also deactivates those with high responsiveness, thus discarding their contributions.

Finally, we can see that the $ne$ value that yields the highest performance varies across different datasets, which validates that the degree of redundancy varies with the task, and thus the pattern of heads that Heart-LoRA should identify and deactivate is distinct across tasks. This confirms the ability of Heart-LoRA to uniquely discover task-specific patterns for different tasks.

\section{Other Ways to Obtain Responsiveness}
\label{alter}
\begin{table}[hb]
\centering
\resizebox{0.47\textwidth}{!}{
\begin{tabular}{l *{4}{c}} % 使用 c 列类型将数字居中对齐
\toprule
Method & Average & Natural & Specialized & Structured \\\hline
\specialrule{0em}{1pt}{1pt}

Full & 68.9 & 75.9 & 83.4 & 47.6 \\
LoRA & 76.4 & 81.5 & 85.2 & 62.6 \\
\rowcolor{lightgray} 
Heart-LoRA$^\dag$ (weight) & 76.7 & 81.6 & 85.8 & 62.5 \\
\rowcolor{lightgray} 
Heart-LoRA$^\dag$ (grad) & 76.7 & 81.6 & 85.5 & 63.0 \\
\bottomrule
\end{tabular}
}
\vspace{0pt}
\caption{Results on VTAB-1K of Heart-LoRA$^\dag$, using L2 norm of weight (denoted as weight) or L2 norm of gradient (denoted as grad) to calculate the responsiveness.}
\label{other-criterion}
\end{table}

Although we utilized Taylor expansion in the section \ref{method} to calculate responsiveness, it is feasible to use other simpler and more straightforward methods as well. Here, we present the results of calculating responsiveness using two criteria: the L2 norm of weight with one parameter or the L2 norm of gradient, as shown in Table \ref{other-criterion}.

From the results, we can observe that using Heart-LoRA$^\dag$, we still manage to surpass LoRA and match Bi-LoRA. Considering this was achieved with a search between $ne$ values of only 1 and 3, it is quite impressive. This outcome stems from the unique adaptation process of Heart-LoRA, which identify the redundant heads and deactivate them during fine-tuning. However, it is noted that this result does not outperform the effectiveness of Heart-LoRA$^\dag$ using Taylor expansion, which underscores the superiority of approach from the section \ref{method}. By considering both weight and gradient simultaneously, it allows for the identification of heads with truly minimal responsiveness, thereby reducing the risk of misidentification.

\section{Comparison to Arbitrary Deactivation}
\label{compare}

We compare the results of the arbitrary deactivation approach adopted in section \ref{arbitrary deactivation} with Heart-LoRA. The results are presented in Table \ref{comparison}. As we can see, although the arbitrary mask performs quite well, exceeding LoRA by roughly $0.5\%$, it still underperforms Heart-LoRA by $0.3\%$, which is notable. This indicates that the performance gain of Heart-LoRA is not only due to the structural advantage of head deactivation but also stems from the unique advantage of using responsiveness to cleverly identify unimportant heads.

While there is improvement when utilizing arbitrary deactivation, we did not utilize the property of head redundancy to the extreme as we were arbitrarily selecting heads to deactivate and ignore the fact that \textit{head-level responsiveness varies across tasks}, as discussed in section \ref{vary}. One option to maximize the performance is to study the attention patterns for each downstream task manually. However, manually studying the patterns is not feasible when dealing with a significant amount of downstream tasks  for it's huge computational cost. This leads us to seek methods capable of efficiently identifying deactivation candidates in the study of PETL without the need for manual investigation, and this brings us to our proposed Heart-LoRA which utilize task-specific responsiveness to cleverly recognize unimportant heads to deactivate. 

\begin{table}[ht]
\centering
\resizebox{0.47\textwidth}{!}{
\begin{tabular}{l *{4}{c}} % 使用 c 列类型将数字居中对齐
\toprule
Method & Average & Natural & Specialized & Structured \\\hline
\specialrule{0em}{1pt}{1pt}

Full & 68.9 & 75.9 & 83.4 & 47.6 \\
LoRA & 76.4 & 81.5 & 85.2 & 62.6 \\
Front & 76.9 & 81.7 & 85.8 & 63.3 \\
\rowcolor{gg}
Heart-LoRA & 77.2 \textcolor{ForestGreen}{($\uparrow 0.3$)} & 81.8 \textcolor{ForestGreen}{($\uparrow 0.1$)} & 86.2 \textcolor{ForestGreen}{($\uparrow 0.4$)} & 63.5 \textcolor{ForestGreen}{($\uparrow 0.2$)} \\ \bottomrule
\end{tabular}
}
\vspace{0pt}
\caption{Comparison between using arbitrary deactivation and the original Heart-LoRA. }
\label{comparison}
\end{table}

\section{Experimental details}

\subsection{Actual Implementation}
\label{code}
The core of Heart-LoRA lies in obtaining a task-specific mask, with the actual code implementation in Alg. \ref{alg:code}:

\begin{algorithm}
\caption{Implementation of obtaining the mask using PyTorch. The obtained mask is then used to perform the deativation of heads.}
\label{alg:code}

\begin{minipage}{0.44\textwidth}
\begin{lstlisting}[language=Python]
def get_scores(args, adapter, num_of_heads=12):
    weight = get_weight(adapter)
    grad = get_grad(adapter)
    step = weight.shape[1] // 12
    all_weights = [weight[:, i:i + step] for i in range(num_of_heads)]
    all_grads = [grad[:, i:i + step] for i in range(num_of_heads)]
    scores = [get_score(all_weights[i], all_grads[i]) for i in range(num_of_heads)]
    return scores

def get_mask(args, adapter, mask, ne):
    scores = get_scores(args, adapter, 12)
    scores_tensor = torch.tensor(scores)
    _, indices = torch.topk(scores_tensor, ne, largest=False)
    indices_list = indices.tolist()
    for i in indices_list:
        mask[:, :, i, :, :] = 0
    return mask
\end{lstlisting}
\end{minipage}

\end{algorithm}

\subsection{Backbones}

\begin{table}[ht]

\begin{center}
\setlength{\tabcolsep}{0.3pt}
\scalebox{0.8}{

\begin{tabular}{p{3.8cm}<{\centering}p{2.7cm}<{\centering}p{1.5cm}<{\centering}p{2cm}<{\centering}}
\hline
\specialrule{0em}{1pt}{1pt}
Model&\makecell[c]{Pre-Training\\Dataset}&Size (M)&\makecell[c]{Pre-Trained\\Weights}
\\\hline\specialrule{0em}{1pt}{1pt}
ViT-B/16~\cite{dosovitskiy2020ViT}&ImageNet-21K&85.8&\href{https://storage.googleapis.com/vit_models/imagenet21K/ViT-B_16.npz}{checkpoint}
\\
Swin-B~\cite{liu2021swintransformer}&ImageNet-21K&86.7&\href{https://github.com/SwinTransformer/storage/releases/download/v1.0.0/swin_base_patch4_window7_224_22k.pth}{checkpoint}
\\

\bottomrule
\end{tabular}
}
\end{center}
\caption{\textbf{Pre-Trained backbones.}
}
\label{tab:pt}
\end{table}
\subsection{Implementation Environment}

On a single NVIDIA RTX 4090 GPU with OS being Ubuntu 22.04.3 LTS x86\_64, we conduct all experiments using \href{https://pytorch.org/}{\emph{PyTorch}} and \href{https://rwightman.github.io/pytorch-image-models/}{\emph{timm}} library.

\subsection{Data Augmentation}

\subsubsection{VTAB-1K}
Following \cite{jie2023revisiting}, we simply adjust the image dimensions to $224\times224$.

\subsubsection{Few-shot Learning}
Also following \cite{zhang2024NOAH} and \cite{jie2023revisiting}, we apply color-jitter and RandAugmentation to the training samples. For the validation/test samples, we first resize them to $256\times256$, then center-crop them to $224\times224$, and finally normalize them using the mean and standard deviation of ImageNet.

\subsection{Hyper-parameter}

Following \cite{jie2023revisiting}, we continue to employ a scaling parameter \( s \). Specifically, our search for \( s \) is conducted within the set \{0.01, 0.1, 1.0, 10, 100\}. Other hyper-parameters are in Table \ref{hyper}, which is basically the same as those in \cite{jie2023revisiting} and \cite{zhang2024NOAH}. More information about the datasets used is provided in Table \ref{dataset}. For parameters $ne$, when the head-level deactivation of Heart-LoRA result in performance degradation, we simply opt for a hyper-parameter setting of $ne=0$. In this scenario, no information is discarded and $P$ is simply all ones. 

\begin{table*}[t]
\begin{center}
\scalebox{0.9}{
\begin{tabular}{llllll}
\toprule[1.5pt]
 & Dataset & \#Classes & Train & Val & Test \\ \midrule
\multirow{19}{*}{VTAB-1k} & CIFAR100 & 100 & \multirow{19}{*}{800/1,000} & \multirow{19}{*}{200} & 10,000 \\
 & Caltech101 & 102 & & & 6,084 \\
 & DTD & 47 & & & 1,880 \\
 & Oxford-Flowers102 & 102 & & & 6,149 \\
 & Oxford-Pets & 37 & & & 3,669 \\
 & SVHN & 10 & & & 26,032 \\
 & Sun397 & 397 & & & 21,750 \\
 & Patch Camelyon & 2 & & & 32,768 \\
 & EuroSAT & 10 & & & 5,400 \\
 & Resisc45 & 45 & & & 6,300 \\
 & Retinopathy & 5 & & & 42,670 \\
 & Clevr/count & 8 & & & 15,000 \\
 & Clevr/distance & 6 & & & 15,000 \\
 & DMLab & 6 & & & 22,735 \\
 & KITTI-Dist & 4 & & & 711 \\
 & dSprites/location & 16 & & & 73,728 \\
 & dSprites/orientation & 16 & & & 73,728 \\
 & SmallNORB/azimuth & 18 & & & 12,150 \\
 & SmallNORB/elevation & 18 & & & 12,150 \\ \midrule
\multirow{5}{*}{Few-shot} & Food-101 & 101 & \multirow{5}{*}{(1/2/4/8/16)*(\#Classes)} & 20,200 & 30,300 \\
 & Stanford Cars & 196 & & 1,635 & 8,041 \\
 & Oxford-Flowers102 & 102 & & 1,633 & 2,463 \\
 & FGVC-Aircraft & 100 & & 3,333 & 3,333 \\
 & Oxford-Pets & 37 & & 736 & 3,669 \\ \bottomrule[1.5pt]
\end{tabular}
}
\end{center}
\caption{\textbf{Dataset Details.}
}
\label{dataset}
\end{table*}

\begin{table*}[t]

\begin{center}
\scalebox{0.9}{
\begin{tabular}{cccccccc}
\toprule[1.5pt]
 &optimizer&batch size&learning rate&weight decay&\# epochs&lr decay&\# warm-up epochs\\ \midrule
 VTAB-1K&AdamW&64&1e-3&1e-4&100&cosine&10\\
 Few-shot learning&AdamW&64&5e-3&1e-4&100&cosine&10\\
 \bottomrule[1.5pt]
\end{tabular}}\end{center}
\caption{\textbf{Other Hyper-parameters.}
}
\label{hyper}
\end{table*}

\end{document}